\documentclass[final,3p,times]{elsarticle}

\usepackage{mathtools}

\usepackage{amssymb}
\usepackage{subcaption}

\usepackage{amsthm}
\usepackage{amsmath}
\usepackage{enumitem}
\usepackage{booktabs}
\usepackage{multirow}
\usepackage{lscape}
\usepackage{tabu}
\usepackage{threeparttable}
\usepackage{natbib}

\usepackage{tikz}
\usetikzlibrary{shapes.geometric, arrows}
\tikzstyle{startstop} = [rectangle, rounded corners, minimum width=3cm, minimum height=1cm,text centered, draw=black, fill=red!30]
\tikzstyle{input} = [rectangle, minimum width=1cm, minimum height=1cm, text centered, draw=black, fill=blue!5, text width=2cm]
\tikzstyle{output} = [rectangle, minimum width=1cm, minimum height=1cm, text centered, draw=black, fill=blue!20, text width=3cm]
\tikzstyle{NN} = [trapezium, trapezium left angle=60,trapezium right angle=60,, minimum width=4.5cm, minimum height=1cm, text centered, draw=black, fill=gray!40]

\tikzstyle{process} = [rectangle, rounded corners, minimum width=4.5cm, minimum height=1cm, text centered, draw=black, fill=gray!20]
\tikzstyle{decision} = [diamond, minimum width=3cm, minimum height=1cm, text centered, draw=black, fill=green!5]
\tikzstyle{arrow} = [thick,->,>=stealth]
\tikzstyle{arrow2} = [dashed,->,>=stealth]

\usepackage[pagewise]{lineno}

\usepackage{setspace}

\journal{Accident Analysis \& Prevention Journal}

\begin{document}

\begin{doublespacing}
\begin{frontmatter}
    
    
    
    \title{\textbf{CGAN-EB: A Non-parametric Empirical Bayes Method for Crash Hotspot Identification Using Conditional Generative Adversarial Networks: \\ A Simulated Crash Data Study\\}}
    
    
    \author{Mohammad Zarei $^1$}
    \author{Bruce Hellinga $^2$}
    \author{Pedram Izadpanah $^3$}

    \address{$^1$ Ph.D. Candidate, Department of Civil and Environmental Engineering, University of Waterloo, 200 University Ave., Waterloo, ON N2L3G1, Canada (corresponding author). E-mail: mzarei@uwaterloo.ca}
    \address{$^2$ Professor, Department of Civil and Environmental Engineering, University of Waterloo, 200 University Ave., Waterloo, ON N2L3G1, Canada. E-mail: bruce.hellinga@uwaterloo.ca }
    \address{$^3$ Adjunct Assistant Professor, Department of Civil and Environmental Engineering, University of Waterloo, 200 University Ave., Waterloo, ON N2L3G1, Canada. E-mail: pedram.izadpanah@uwaterloo.ca}

    \begin{abstract}
        In this paper, a new non-parametric empirical Bayes approach called CGAN-EB is proposed for approximating empirical Bayes (EB) estimates in traffic locations (e.g., road segments) which benefits from the modeling advantages of deep neural networks, and its performance is compared in a simulation study with the traditional approach based on negative binomial model (NB-EB). The NB-EB uses negative binomial model in order to model the crash data and is the most common approach in practice. To model the crash data in the proposed CGAN-EB, conditional generative adversarial network is used, which is a powerful deep neural network based method that can model any types of distributions. A number of simulation experiments are designed and conducted to evaluate the CGAN-EB performance in different conditions and compare it with the NB-EB. The results show that CGAN-EB performs as well as NB-EB when conditions favor the NB-EB model (i.e. data conform to the assumptions of the NB model) and outperforms NB-EB in experiments reflecting conditions frequently encountered in practice, specifically low sample means, and when crash frequency does not follow a log-linear relationship with covariates.

    \end{abstract}
    

\end{frontmatter}
\end{doublespacing}


\section{Introduction and background}
\label{section:intro}

When it comes to highway safety management \cite{HSM}, identifying crash hotspots (also known as network screening) is a crucial part of the process. Traffic locations (e.g., intersections, road segments) are examined and ranked depending on some crash risk measures \cite{yu2014comparative,thakali2015identification,li2021toward}. For example, in the empirical Bayes (EB) method the observed crash counts of a site is combined with the predicted safety (number of crashes) of similar sites to generate the EB estimate \cite{hauer1997observational}. The prediction is typically derived from a safety performance function (SPF) that relates crash frequency as a function of variables such as road characteristics and traffic exposure.

SPFs are crash predictive models that are calibrated to crash data, usually using a parametric modeling approach. A parametric model that is widely used in practice is the Negative Binomial (NB) model, in large part because it can be readily implemented together with the EB method and it can handle over-dispersion of crash data. However, other parametric models have been proposed that can also account for over-dispersion and be implemented within the EB framework, including Poisson-lognormal \cite{miranda2005possionlognormal}, Sichel model \cite{zou2013Sichel}, and NB-Lindley \cite{geedipally2012NB-Lindley}. Each of these models may have better results in different data sets. 

There are several challenges to developing SPFs using parametric modeling. Selecting the best parametric model and establishing a functional form that best fits the given crash data can be technically challenging and time/effort intensive and the choice can have a significant influence on the outcome of network screening \cite{spf_form1,spf_form2}. One alternative which avoids these challenges is to use non-parametric or semi-non-parametric modeling approaches. For instance in \cite{ye2018semi}, it has been shown that a semi-non-parametric Poisson (SNP) model performs better than NB model when calibrated to simulated data sets with various distribution of error terms. The NB model is found to substantially overestimate the effect of lane width on crash frequency reduction relative to the SNP model based on more robust estimation of unobserved heterogeneity.

Another option to deal with the aforementioned challenges of parametric models is using fully non-parametric or data-driven models such as deep neural networks (DNN). Several studies have shown that such models have better fitting and predictive performance than parametric ones \cite{dong2018NN,zeng2016NN,huang2016NN, pan2017NN, karlaftis2011statisticalvsNN, singh2020deepNN}. These computational models are able to extract inherent features in the data minimizing the efforts required for feature selecting and feature engineering that is essential for parametric models. They have also demonstrated superior performance in dealing with multi-colinearity, which is the non-independence of predictor variables \cite{obite2020multicollinearity}. Moreover, significant progress has been made to address the criticism that DNN-based models are black-boxes \cite{buhrmester2019analysis, samek2019explainable}. In one of the recent works, a novel method has been proposed that can be used to describe the inner working of deep learning models \cite{chen2021novel}. The method includes visualization and feature importance criteria over the input and hidden layers of the network that can be used for data/feature importance analysis.

Non-parametric approaches, such as DNN models, have been applied to different traffic safety problems to overcome these limitations. Some of the recent applications of deep learning models in crash data analysis include a crash count model with an embedded multivariate negative binomial model \cite{dong2018NN}, developing a global safety performance function \cite{pan2017NN}, real time crash predictions \cite{theofilatos2019realtime, li2020real}, pedestrian near-accident detection \cite{zhang2020pedstrian}, crash severity prediction \cite{rezapour2020severity,zheng2019severity}, and crash data augmentation \cite{islam2021vae, cai2020real}. However, using a DNN based model for the purpose of hotspot identification and integrating it with the EB method is quite rare. Hence, there is a need for an EB estimate approach that takes advantage of DNN models and does not have the limitations of parametric approaches.

One of the recently developed deep learning models is generative adversarial network, or GAN for short, that can implicitly model any kind of data distribution \cite{goodfellow2014GAN}. GAN, which has achieved tremendous success in many fields (e.g., image/video synthesis/manipulation, natural language processing, classification) in recent years \cite{goodfellow2014GAN,goodfellow2016GANtutorial,creswell2018ganreview}, consists of simultaneously training two deep neural networks, a generator which produces synthetic samples that mimic the characteristics (i.e. distribution) of the real (observed) data, and a discriminator which tries to distinguish between the synthetic samples coming from the generator and the real samples from the original data set. In a conditional GAN \cite{mirza2014cgan}, referred to as CGAN, both the generator and the discriminator are conditioned on some data which could be a class label or a feature vector if we wish to use it for regression purposes. GAN and its variants have been rarely used in crash data analysis. Our review of the literature indicates very little application of GAN and its variants to crash data analysis problems. We did observe the use of GAN for generating traffic data related to crashes to address data imbalance in real-time crash prediction \cite{cai2020real}. In other transportation areas, GAN have been recently used for network traffic prediction \cite{zhang2019trafficgan} and traffic flow data imputations \cite{chen2019traffic}.

The goal of this paper is twofold: to propose an EB estimation method based on CGAN model and to evaluate the performance of the proposed CGAN-EB with the traditional approach referred to as NB-EB. To this end, a simulation study has been conducted in order to simulate different crash data sets with known regression parameters describing the mean and the dispersion level. For each simulated data set, the accuracy of EB estimates obtained from both NB-EB and CGAN-EB methods are compared. A simulation environment has been frequently used for traffic safety studies \cite{young2014simulation,park2014finite,ye2018semi} specifically because it provides the following two benefits over using empirical data; (a) a range of specified  conditions (e.g. modifying dispersion, mean, data size) can be evaluated, and (b) the true crash risk values at each location are known \cite{zou2015modeling}. The details regarding simulation steps, calculation of EB estimates and evaluation process are provided in the following sections.

\section{Empirical Bayes estimation for crash risk}
\label{section:EB}

\subsection{Parametric approach: NB-EB}
\label{section: NB-EB}

As proposed by Hauer \cite{hauer1997observational}, given $K$ as the observed number of crashes and $k$ as the expected number of crashes, the EB estimator of $k$ can be calculated as follows:
\begin{equation}
    \label{Eq:EB}
    E(k|K) = w \times E(k) + (1-w)\times K \approx w \times E(K) + (1-w)\times K
\end{equation}

\noindent where the weight $w$ is shown to be a function of the mean and variance of $k$ and is always a number between 0 and 1:
\begin{equation}
    \label{Eq:w}
    w = \frac{E(k)}{E(k) + Var(k)} 
\end{equation}

\noindent In the above equations, if $k$ is gamma distributed, then the resulting $K$ follows an NB distribution (Eq. \ref{Eq:NB}). As a result, the NB-EB estimate can be derived as follows:

\begin{equation}
    \label{Eq:NB}
    f(y|\mu, \alpha) = \frac{\Gamma(y+1/\alpha)}{\Gamma(1/\alpha)\Gamma(y+1)}\left( \frac{\alpha \mu}{1+\alpha \mu}\right)\left( \frac{1}{1+\alpha \mu}\right)
\end{equation}

\begin{equation}
    \label{Eq:NB_EB}
    EB^{NB} = w \times \mu + (1-w) \times y
\end{equation}
\begin{equation}
    w = \frac{1}{1+\alpha \mu}
\end{equation}

\noindent where  $y$ is the observed number of crashes per year, $\alpha$ is  dispersion parameter, and $\mu$ is the number of crashes predicted by the NB model. Note that the mean and the variance of $y$ are $E[y] = \mu$ and $var(y) = \mu + \alpha \mu^2$ respectively. If $\alpha \rightarrow 0$, the crash variance equals the crash mean and the NB distribution converges to the Poisson distribution. Essentially, the EB method employs a weighted average of the observed and predicted crash counts, so that if the variance of prediction is high, lower weights are assigned to the prediction, and vice versa \cite{hauer2002tutorial}.

\subsection{Non-parametric approach: CGAN-EB}
\label{section: CGAN-EB}

\begin{figure}[tbp]
    \centering
    
    \begin{tikzpicture}[auto, node distance=4cm,>=latex']

        \node (G) [NN, yshift = 0cm] {\textbf{Generator}};
        \node (G_out) [output, above of = G, xshift = 0cm, yshift=-2cm] {\footnotesize $G(X, z)=\hat{y}$};
        
        \node (noise) [input, left of = G, yshift = 0.6cm] {\footnotesize Noise ($z$)};
        \node (X) [input, left of = G, yshift = -0.6cm] {\footnotesize $X$};
        
        \node (D) [NN, above of= G_out, yshift=-2cm] {\textbf{Discriminator}};
        
        \node (real) [input, above of=D, yshift=-2.5cm, text width = 4cm] {\footnotesize Real  data $(X, y)$};
        
        \node (D_out_real) [output, left of= D, yshift = -1.5cm, xshift = -1cm, text width = 4cm] {\footnotesize Real Loss: $\mathbf{L(}D(X,y), \mathbf{1})$};
        \node (D_out_fake) [output, right of= D, yshift = -1.5cm, xshift = 1cm, text width = 4cm] {\footnotesize Fake Loss: $\mathbf{L(}D(X,\hat{y}), \mathbf{0})$};
        
        \draw [arrow] (noise) -- (G);
        \draw [arrow] (X) -- (G);
        \draw [arrow] (G) -- (G_out);
        \draw [arrow] (G_out) -- (D);
        \draw [arrow] (real) -- (D);
        \draw [arrow] (D) -- (D_out_real);
        \draw [arrow] (D) -- (D_out_fake);
        \draw [arrow2] (D_out_fake) |- (G);
        \draw [arrow2] (D_out_fake) |- (D);
        \draw [arrow2] (D_out_real) |- (D);
 
    \end{tikzpicture}
    
    \caption{CGAN training structure ($X$ is feature vector, $y$ is the dependent variable, $z$ is a noise value from a normal distribution $N(0,1)$)}
    \label{fig:CGAN}
\end{figure}
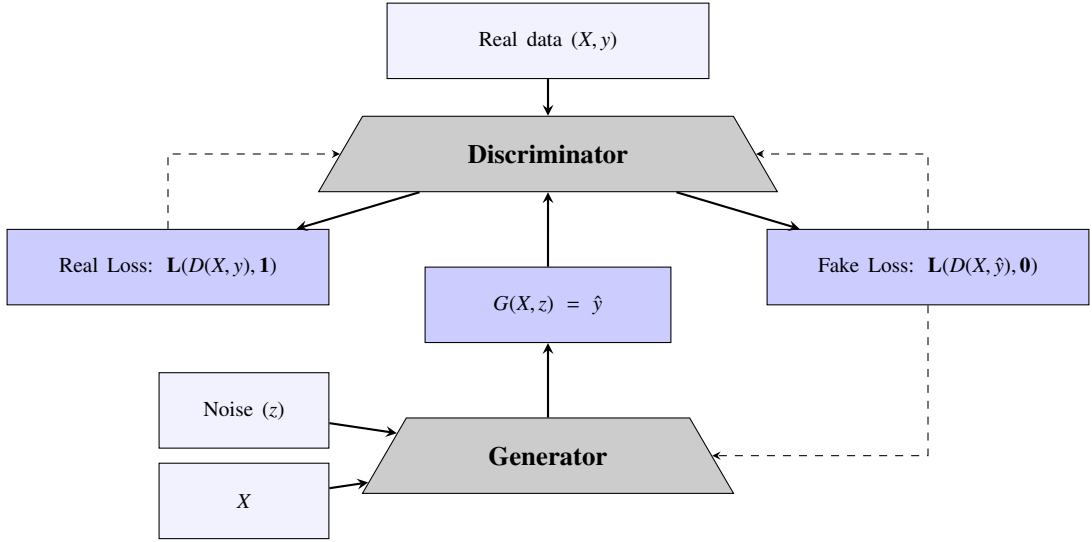

This section describes the training process of a CGAN model and proposes a method to combine it with the EB method. Figure \ref{fig:CGAN} shows the training steps for a CGAN. It begins by assigning two sets of random weight values to both the discriminator and generator neural networks. 

Next, real loss value which shows the ability of the discriminator to recognize the real instances (i.e. $y$) is calculated based on $D(X, y)$, unit vector (i.e. $\mathbf{1}$) and a loss function (e.g. binary cross entropy). Note that $D(X, y) \in [0,1]$ is the output of the discriminator using $(X, y)$ as input and $G(X,z)=\hat{y}$ is the output of generator using $(X,z)$ as input. 

Fake loss value which shows the ability of the discriminator to recognize fake instances (i.e. $\hat{y}$) is calculated based on $D(X, \hat{y})$, zero vector (i.e. $\mathbf{0}$) and a loss function (e.g. binary cross entropy). Here $D(X,\hat{y})$ is the output of the discriminator using $(X,\hat{y})$ as input. Then, the weights of the discriminator are updated based on the objective to minimize total loss (i.e. real loss + fake loss) and the weights of the generator will be updated based on its objective to maximize fake loss. 

This cycle continues until the stop condition (e.g. maximum number of epochs) is met. Both fake and real loss converge to 0.5 in an ideal situation, indicating that it is impossible to discriminate between real and synthetic data because they are samples from the same distribution  \cite{mirza2014cgan}. In the context of crash prediction models, $X$, $y$ and $\hat{y}$ represent site characteristics, crash count, and the generated crash count by generator respectively. 

The generator mimics the underlying distribution of real data so it can generate samples from the same distribution for a given site and this set of samples is conditioned on its feature vector. The mean of the samples can be then be interpreted as the prediction of the model \cite{aggarwal2019cganreg}.

In order to derive the EB estimates using a CGAN model ($EB^{CGAN}$), we need $E(k)$ and $Var(k)$ (see Eq. \ref{Eq:w}) which can be approximated using the samples (e.g. $m=500$ samples) taken from a trained CGAN model given the feature vector ($X$) of any given site:

\begin{equation}
    \label{Eq:CGAN mean}
    E^{CGAN}(k) \approx \frac{\sum_{j=1}^m CGAN_j(X)}{m}
\end{equation}
\begin{equation}
    \label{Eq:CGAN var}
    Var^{CGAN}(k) \approx \frac{\sum_{j=1}^m (E^{CGAN}(k)-CGAN_j(X))^2}{m-1}
\end{equation}

\noindent where $CGAN_j(X)$ is the $j$-th sample from total $m$ samples obtained from the CGAN model when provided with $X$, the feature vector of a given site, as input. Also, the weight factor $w$ in Eq. \ref{Eq:EB} when using a CGAN model can be defined as follows:

\begin{equation}
\label{Eq:CGAN w}
    w^{\small CGAN} \approx \frac{E^{CGAN}(k)}{E^{CGAN}(k)+ Var^{CGAN}(k)}
\end{equation}

Based on the above equations, $EB^{CGAN}$ can be formulated as:

\begin{equation}
    \label{Eq:EB CGAN}
    EB^{CGAN} = w^{\small CGAN} \times E^{CGAN}(k) + (1-w^{\small CGAN})\times y
\end{equation}

\section{Simulation experiments and evaluation methods}
\label{section: simulation}

The simulation experiments used to evaluate the performance of CGAN-EB versus NB-EB are described in this section. Twelve experiments are designed and presented in Table \ref{Table:Experiments} to investigate the impact of dispersion parameter, sample mean, and  sample size on the performance of CGAN-EB versus NB-EB. In order to simulate conditions that are convincingly similar to conditions in empirical crash data, the parameters of these experiments (i.e. dispersion, sample mean and sample size) are based on the reported parameters related to crash datasets in eight published papers summarized in \cite{zou2015modeling}.

\begin{table}[ht]
\centering
\renewcommand{\arraystretch}{1.2} 
\caption{Experiments }
    \begin{tabular}{ccccc}
    \toprule
              & \multicolumn{2}{c}{Low Dispersion ($\alpha=0.5$)} & \multicolumn{2}{c}{High Dispersion ($\alpha=1.5$)} \\ \cmidrule{2-3} \cmidrule{4-5}
Sample size & Low Mean ($\Bar{y}=1.5$)    & High Mean ($\Bar{y}=12$)    & Low Mean ($\Bar{y}=1.5$)      & High Mean ($\Bar{y}=12$)     \\  \midrule
2000          & E1           & E2           & E3            & E4             \\
1000          & E5           & E6           & E7            & E8             \\
500           & E9           & E10          & E11           & E12    \\ \bottomrule
    \end{tabular}
    
\label{Table:Experiments}
\end{table}

For each of these experiments, five training data sets are randomly generated based on the experiment parameters and using the following steps which have been proposed in \cite{francis2012characterizing} and been used in previous simulation studies \cite{zou2015modeling, ye2018semi}:

\begin{enumerate}
    \item Generate a random feature vector with the size of 4 ($X_1, X_2, X_3, X_4$) from a uniform distribution on $[0, 1]$.
    \item Generate the corresponding count $Y_i$ given that the mean for observation $i$ is gamma distributed with the dispersion parameter $\alpha$ and mean equal to 1:
    \begin{itemize}
        \item[] $Y_i \sim Poisson (\lambda_i)$;
        \item[] $\lambda_i = \exp(\beta_0 + 0.05X_1 - 0.05X_2 + X_3 - X_4 + \epsilon_i)$;
        \item[] $\exp(\epsilon_i) \sim gamma (1,\alpha)$.
    \end{itemize}
    \item Steps (1) and (2) are repeated until the sample size associated with each experiment is reached.
\end{enumerate}

Note that $\epsilon$ represents the unobserved heterogeneity following the log-gamma distribution as assumed in NB model. The sample mean is controlled by $\beta_0$ which is equal to 0.5 for low mean experiments and 2.5 for high mean experiments. It is worth mentioning that these simulation settings are completely consistent with the NB model assumptions regarding the error term distribution and log-linear relationship between dependent and independent variables.

\begin{figure}[ht]
    \centering
    \includegraphics[width=0.8\linewidth]{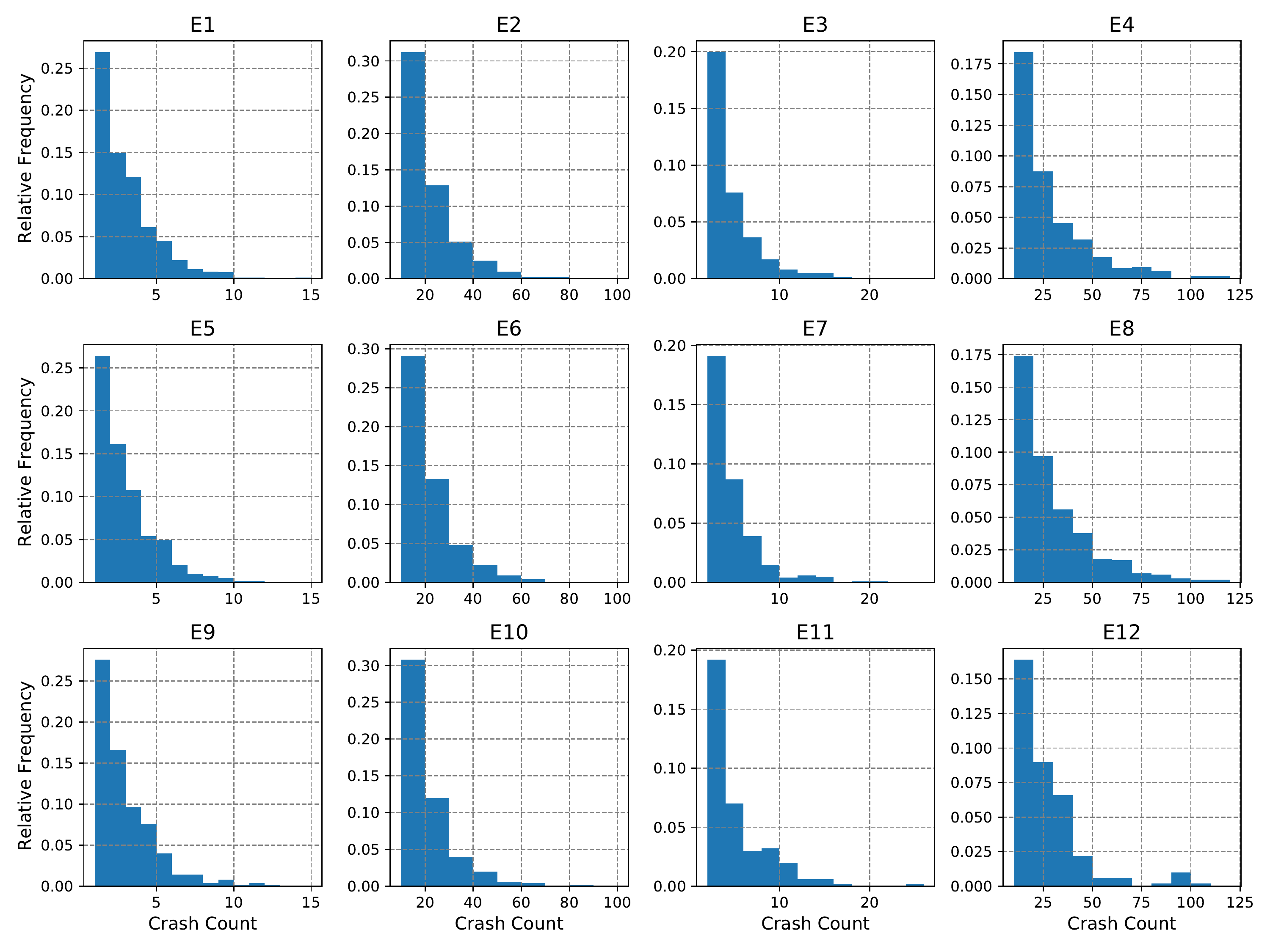}
    \caption{Distribution of simulated crash counts for one data set of each experiment}
    \label{fig:Dists}
\end{figure}

To illustrate the simulated crash data the distribution of simulated crash counts (i.e. $Y_i$s) for one data set of each experiment is presented in Figure \ref{fig:Dists}.

\begin{figure}
\small
  \centering
  \begin{subfigure}{0.48\textwidth}
    \begin{tikzpicture}[auto, node distance=4cm,>=latex']
        
        \node (concat) [process, minimum height = 0.6cm] {ConcatLayer(200)};
        \node (X_out) [NN, above of = concat, xshift=-2cm, yshift=-3cm, minimum height = 0.6cm, minimum width = 2cm] {Dense(100, ELU)};
        \node (X) [input, minimum height = 0.6cm,above of = X_out, xshift=0cm, yshift=-3cm] {$X$};
        
        \node (noise_out) [NN, above of = concat, xshift=2cm, yshift=-3cm, minimum height = 0.6cm, minimum width = 2cm] {Dense(100, ELU)};
        \node (noise) [input, minimum height = 0.6cm, above of = noise_out, xshift=0cm, yshift=-3cm] {$z$};
        
        \node (l1) [NN, below of = concat, xshift=0cm, yshift=3cm, minimum height = 0.6cm] {Dense(40, ELU)};
        \node (l2) [NN, below of = l1, xshift=0cm, yshift=3cm, minimum height = 0.6cm] {Dense(40, ELU)};
        \node (l3) [NN, below of = l2, xshift=0cm, yshift=3cm, minimum height = 0.6cm] {Dense(40, ELU)};
        \node (l4) [NN, below of = l3, xshift=0cm, yshift=3cm, minimum height = 0.6cm] {Dense(1, ReLU)};

        \draw [arrow] (X) -- (X_out);
        \draw [arrow] (noise) -- (noise_out);
        \draw [arrow] (X_out) -- (concat);
        \draw [arrow] (noise_out) -- (concat);
        \draw [arrow] (concat) -- (l1);
        \draw [arrow] (l1) -- (l2);
        \draw [arrow] (l2) -- (l3);
        \draw [arrow] (l3) -- (l4);

    \end{tikzpicture}
    
    \caption{Generator architecture}
    \label{fig:Generator architecture}
  \end{subfigure}
  \begin{subfigure}{0.48\textwidth}
    \begin{tikzpicture}[auto, node distance=4cm,>=latex']
        
        \node (concat) [process, minimum height = 0.6cm] {ConcatLayer(200)};
        \node (X_out) [NN, above of = concat, xshift=-2cm, yshift=-3cm, minimum height = 0.6cm, minimum width = 2cm] {Dense(100, ELU)};
        \node (X) [input, minimum height = 0.6cm, above of = X_out, xshift=0cm, yshift=-3cm] {$X$};
        
        \node (y_out) [NN, above of = concat, xshift=2cm, yshift=-3cm, minimum height = 0.6cm, minimum width = 2cm] {Dense(100, ELU)};
        \node (y) [input, minimum height = 0.6cm, above of = noise_out, xshift=0cm, yshift=-3cm] {$y$};
        
        \node (l1) [NN, below of = concat, xshift=0cm, yshift=3cm, minimum height = 0.6cm] {Dense(40, ELU)};
        \node (l2) [NN, below of = l1, xshift=0cm, yshift=3cm, minimum height = 0.6cm] {Dense(40, ELU)};
        \node (l3) [NN, below of = l2, xshift=0cm, yshift=3cm, minimum height = 0.6cm] {Dense(1, Sigmoid)};

        \draw [arrow] (X) -- (X_out);
        \draw [arrow] (y) -- (y_out);
        \draw [arrow] (X_out) -- (concat);
        \draw [arrow] (y_out) -- (concat);
        \draw [arrow] (concat) -- (l1);
        \draw [arrow] (l1) -- (l2);
        \draw [arrow] (l2) -- (l3);

    \end{tikzpicture}
    
    \caption{Discriminator architecture}
    \label{fig:Discriminator architecture}
  \end{subfigure}

\caption{Network architectures. The input layer of generator includes normalized feature vector with size of 8 and a noise value ($z\sim N(0,1)$), and input layer of discriminator includes same feature vector and crash count (i.e. $y$)} 
\label{fig:architecture}
\end{figure}
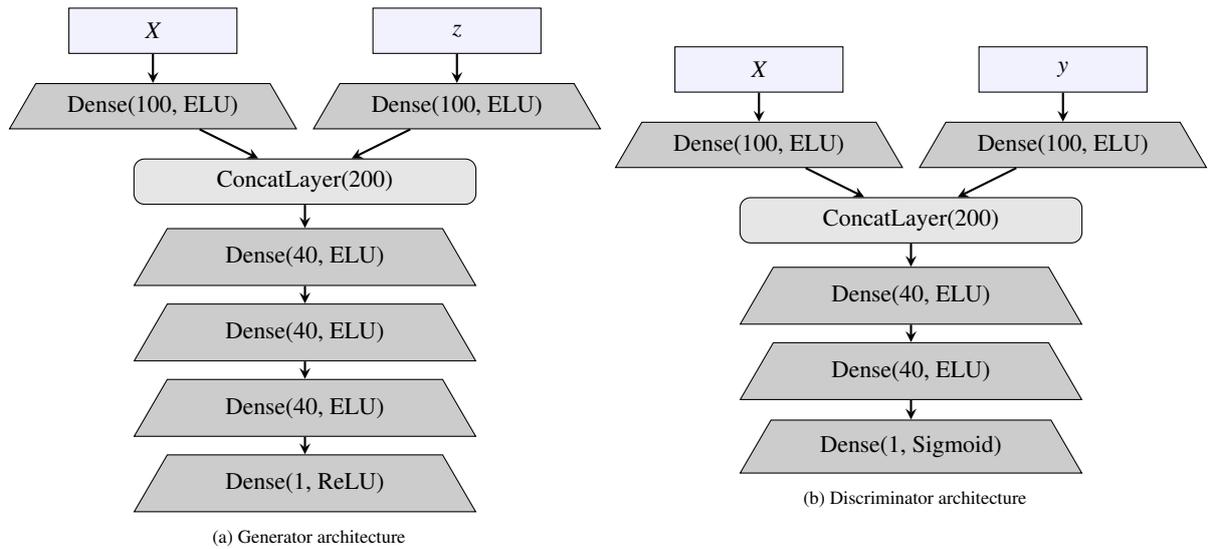

After simulating all data sets for each experiment, we have trained both NB-EB and CGAN-EB methods using each training data set. The NB models have been developed using StatsModels \cite{seabold2010statsmodels}, a Python-based module providing different classes and functions for statistical modeling and analysis. The dispersion parameter (i.e. $\alpha$) is determined using auxiliary Ordinary Least Square (OLS) regression without constant \cite{aux_OLS}. The CGAN models for this study have been developed using Keras \cite{chollet2018keras}, an open-source deep neural network library written in Python. The architectures of the generator and discriminator are presented in Figure \ref{fig:architecture}. These architectures are designed based on suggested architectures in \cite{aggarwal2019cganreg} for using CGAN as a regression model. \textit{DenseLayer(n)} in Figure \ref{fig:architecture} is a regular deeply connected neural network layer with $n$ nodes and \textit{ConcatLayer(n)} concatenates a list of inputs. The model configuration parameters are set as follows:

\begin{itemize} [itemsep=0pt,parsep=0pt, topsep=0pt, partopsep=0pt]
    \item Activation functions: Exponential Linear Unit (ELU), Rectified Linear Unit (ReLU), and Sigmoid \cite{sharma2017activation}
    \item Optimizer: Adam \cite{pedamonti2018comparison}
    \item Number of epochs: 500
    \item Batch size: 100
    \item Learning rate (both generator and discriminator): 0.001
    \item Learning rate decay (generator): 0.001
    \item Learning rate decay (discriminator): 0.0
\end{itemize}

In order to compare CGAN-EB versus NB-EB in terms of their performance as crash hotspot identification methods, two error measures, false identification (FI) test and Poison mean difference (PMD) test proposed in \cite{cheng2008test}, have been employed. These tests have been specifically used for comparing hotspot identification methods in a simulation environment where the truth is known. The FI test calculates the number of sites that are erroneously categorised as hotspots, and the PMD test is the mean absolute difference of the true Poisson means for true hotspots and the suggested hotspots by a method. Because we are comparing a variety of conditions and data sets in this simulation study, a normalised version of these tests is proposed and used. In numerical form, the normalized FI and PMD tests are as follows:

\begin{equation}
\label{eq:FI}
    FI_m = \frac{|\{h_{r=1}, h_{r=2}, ... , h_{r=R}\} - \{x_{r=1}, x_{r=2}, ... , x_{r=R}\}_m |}{|\{h_{r=1}, h_{r=2}, ... , h_{r=R}\}|} 
\end{equation}

\begin{equation}
\label{eq:PMD}
    PMD_m = \frac{\sum \lambda_h - \sum \lambda_x}{\sum \lambda_h} 
\end{equation}

For FI equation, $x_{r=1}, x_{r=2}, \ldots , x_{r=R}$ are the sites suggested as hotspots by the method $m$ that are ranked $1; 2;\ldots ; R$ respectively, and $h_{r=1}, h_{r=2}, \ldots  , h_{r=R}$ are the true hotspots that are ranked based on the true Poisson means. $R$ is the rank threshold that is used as a cut-off in the hotspot identification. For the PMD equation, $\sum \lambda_h$ is the sum of true Poisson means for true hotspots and $\sum \lambda_x$ is the sum of true Poisson means for the sites suggested as hotspots by method $m$. Both the FI and PMD tests have a minimum value of zero (when the method discovers all true hotspots) and a maximum value of 1 (when none of the true hotspots are detected. Note that for each experiment, five pairs of NB and CGAN models are trained using five different training data sets and each model was evaluated using five separate simulated test data sets in order to avoid potential over-fitting of models.  This produced 25 test replications for each experiment.

\section{Results and Discussion}
\label{section:results}
\begin{figure}[h]
    \centering
    \includegraphics[width=0.9\linewidth]{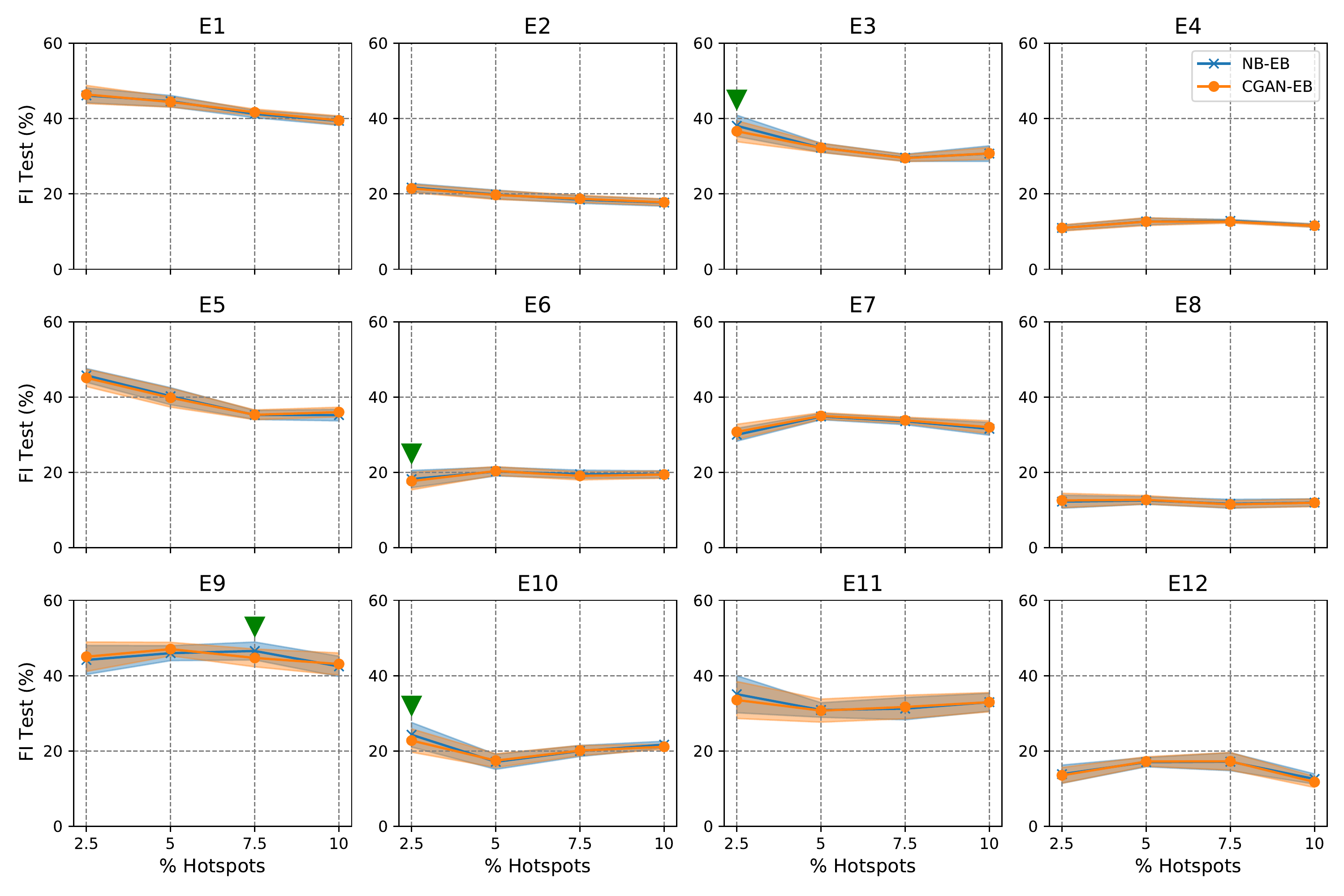}
    \caption{FI test results for the twelve experiments}
    \label{fig:FI}
\end{figure}

\begin{figure}[h]
    \centering
    \includegraphics[width=0.9\linewidth]{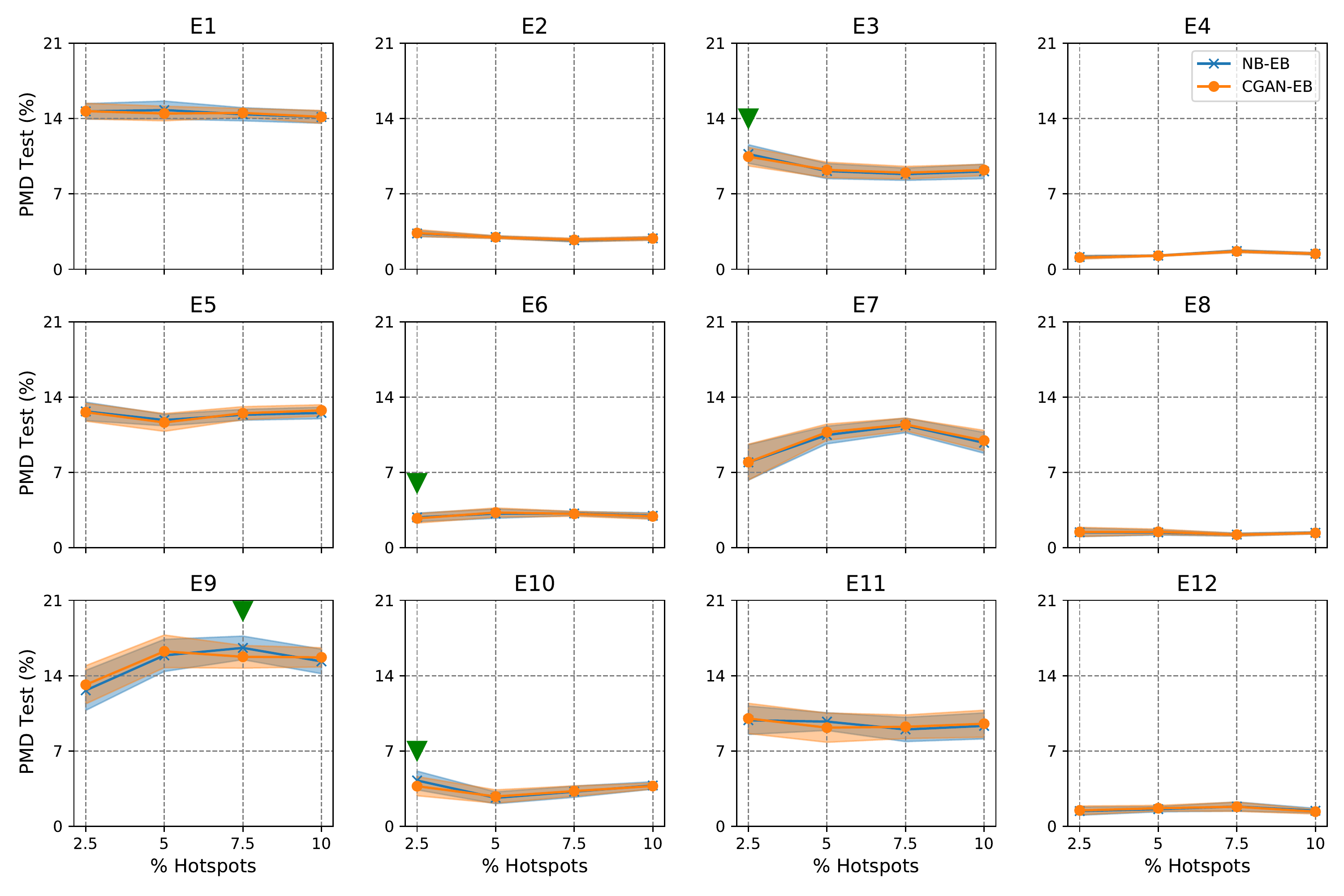}
    \caption{PMD test results for the twelve experiments}
    \label{fig:PMD}
\end{figure}

The goal of the twelve experiments presented in Table \ref{Table:Experiments} is to compare CGAN-EB with NB-EB in terms of hotspot identification performance under different data sizes, sample mean and dispersion parameters. This comparison for each experiment is performed using FI and PMD tests for 2.5\%, 5\%, 7.5\%, and 10\% top hotspots. Since there are 25 replications for each experiment, there are 25 test results for each set of hotspots. The average of these 25 test results for each method (i.e. CGAN-EB and NB-EB), and the corresponding 95\% confidence intervals (using t-distribution with DOF = 24) are presented in Figure \ref{fig:FI} for FI test and in Figure \ref{fig:PMD} for PMD test.

Figure \ref{fig:FI} and Figure \ref{fig:PMD} show that the average values and the interval ranges for FI and PMD tests are lower in the experiments with higher sample mean (i.e. E2, E6, E10, E4, E8, E12) in comparing to the experiments with lower sample mean (i.e. E1, E5, E9, E3, E7, E11). In addition, FI and PMD test results have a general increasing trend toward experiments with lower sample sizes (i.e. E9, E10, E11, E12). These observations are expected as it has been shown that the crash data characterized by a low sample mean combined with a small sample size can seriously affect the estimation of the dispersion parameter \cite{lord2006modeling}.

Regarding CGAN-EB versus NB-EB comparison, we performed a paired sample t-test (DOF = 24, p-value = 0.05) for each hotspot percentage in each experiment (44 cases) for FI and PMD results. The outcomes indicate that the difference between FI and PMD results of the two methods are statistically significant in 4 cases for FI and 4 cases for PMD (shown in figures by green triangles) in which the average value of FI and PMD results for CGAN-EB are lower than NB-EB. Consequently, we can conclude that for these cases, the CGAN-EB performs better than NB-EB and for the remaining cases, the CGAN-EB method performance is no different than the NB-EB method.

Recall that for these simulation experiments, all simulation settings were selected to be completely consistent with the assumptions of the NB-EB method, and despite this, the performance of CGAN-EB was at least as good  as NB-EB. As a result, it is expected that CGAN-EB performance relative to the NB-EB performance improves when the simulation settings are changed to become less consistent with the NB assumptions. To that purpose, we ran four more experiments (F5, F6, F7, F8 referred to as F-experiments) using the similar simulation settings as E5, E6, E7, E8 (referred to as E-experiments) for comparison purposes, but with a log-nonlinear functional form instead of a log-linear functional form at step 2 of the simulation in Section \ref{section: simulation}:

\begin{equation}
    \lambda_i = \exp(\beta_0 + 0.05 X_1^{0.5} - 0.05 X_2^{0.5} + X_3^2 - X_1 X_4 + \epsilon_i)
\end{equation}

\begin{figure}[]
    \centering
    \includegraphics[width=0.9\linewidth]{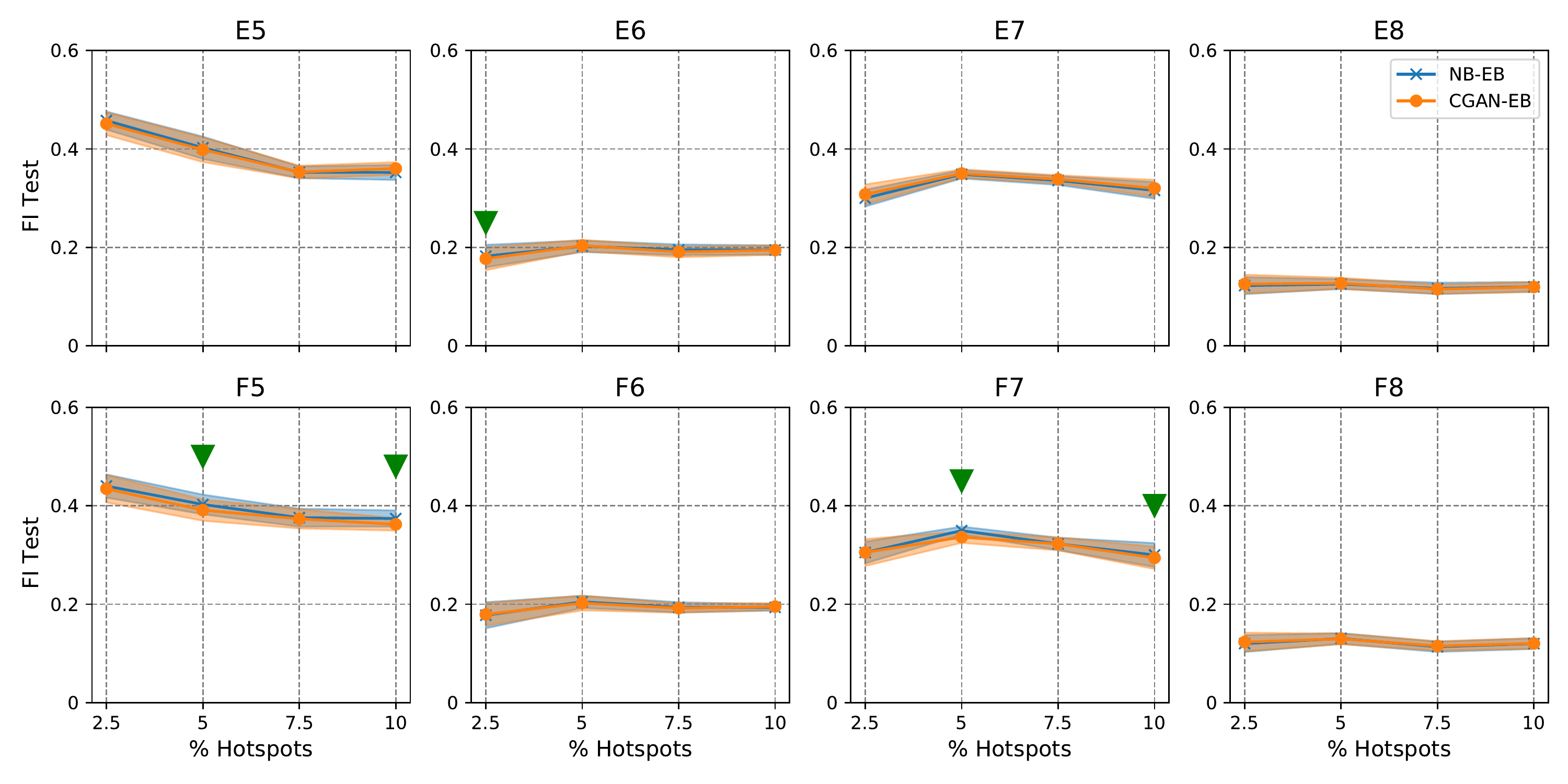}
    \caption{FI test results for E-experiments versus F-experiments}
    \label{fig:FI nonlinear}
\end{figure}

\begin{figure}[]
    \centering
    \includegraphics[width=0.9\linewidth]{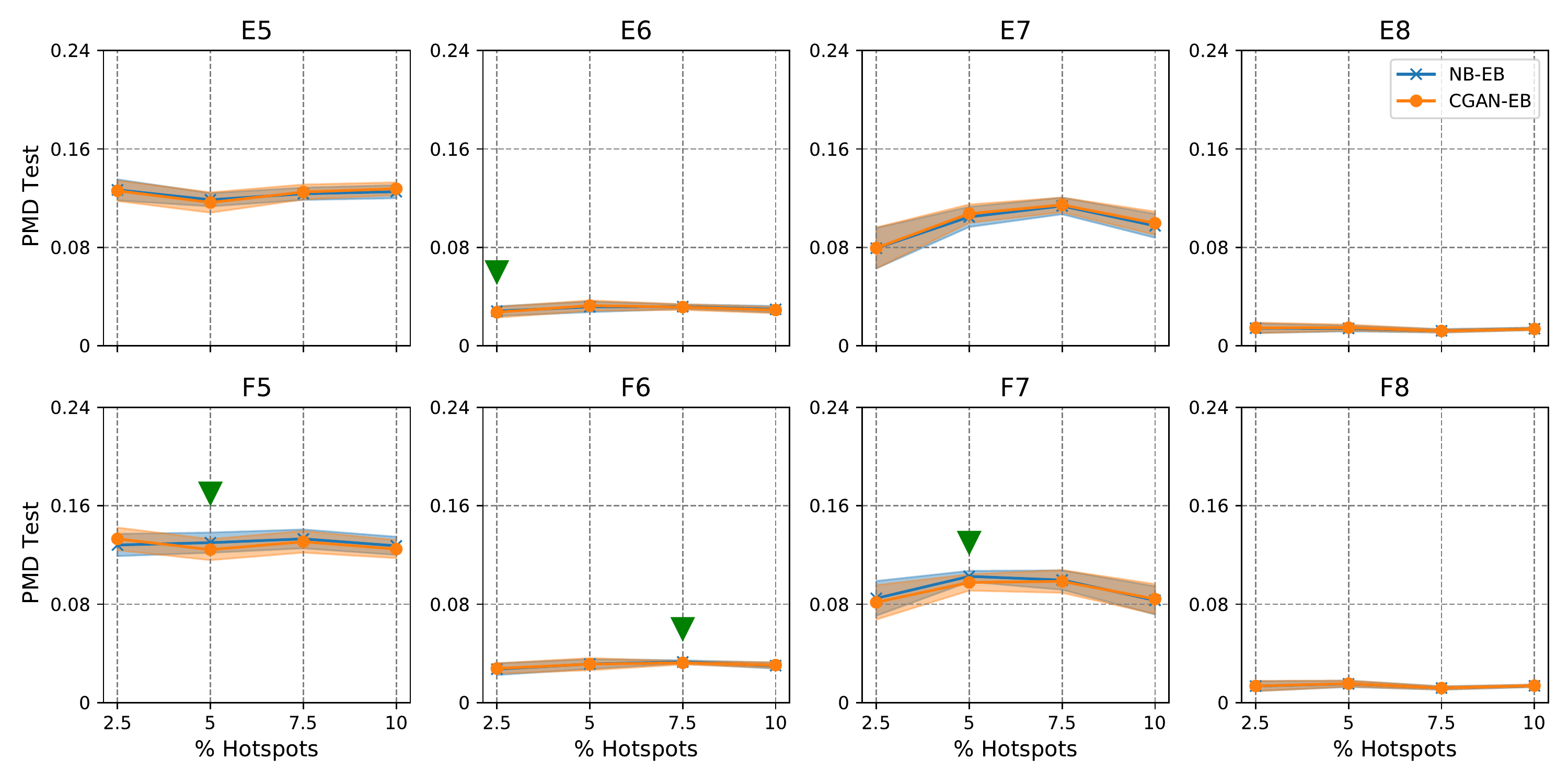}
    \caption{PMD test results for E-experiments versus F-experiments}
    \label{fig:PMD nonlinear}
\end{figure}

Using the same FI and PMD tests, the performance of CGAN-EB and NB-EB are compared within each experiment. The results, which are presented in Figure \ref{fig:FI nonlinear} and Figure \ref{fig:PMD nonlinear}, indicate that more significant improvements are achieved by CGAN-EB over NB-EB in these F-experiments. 

Considering FI test, the paired sample t-test showed that the CGAN-EB model performed better than the NB-EB model for 4 out of 16 F-experiment cases (denoted by green triangles in the figures), and only  1 out of 16 E-experiments cases. In all other cases, the differences of the two methods were not statistically significant. For the PMD test, the CGAN-EB model performed better than the NB-EB model for 3 out of 16 cases in F-experiments cases and 1 out of of 16 cases in E-experiments. In addition, it can be seen that as expected the improved performance of CGAN-EB over NB-EB is generally more notable in experiments with lower sample means (i.e. F5 and F7). The percentage improvement in FI and PMD tests for CGAN-EB in E-experiments can reach up to 3\% (in E6) and up to 5\% (in F7) in F-experiments.

\begin{figure}[]
    \centering
    \includegraphics[width=0.9\linewidth]{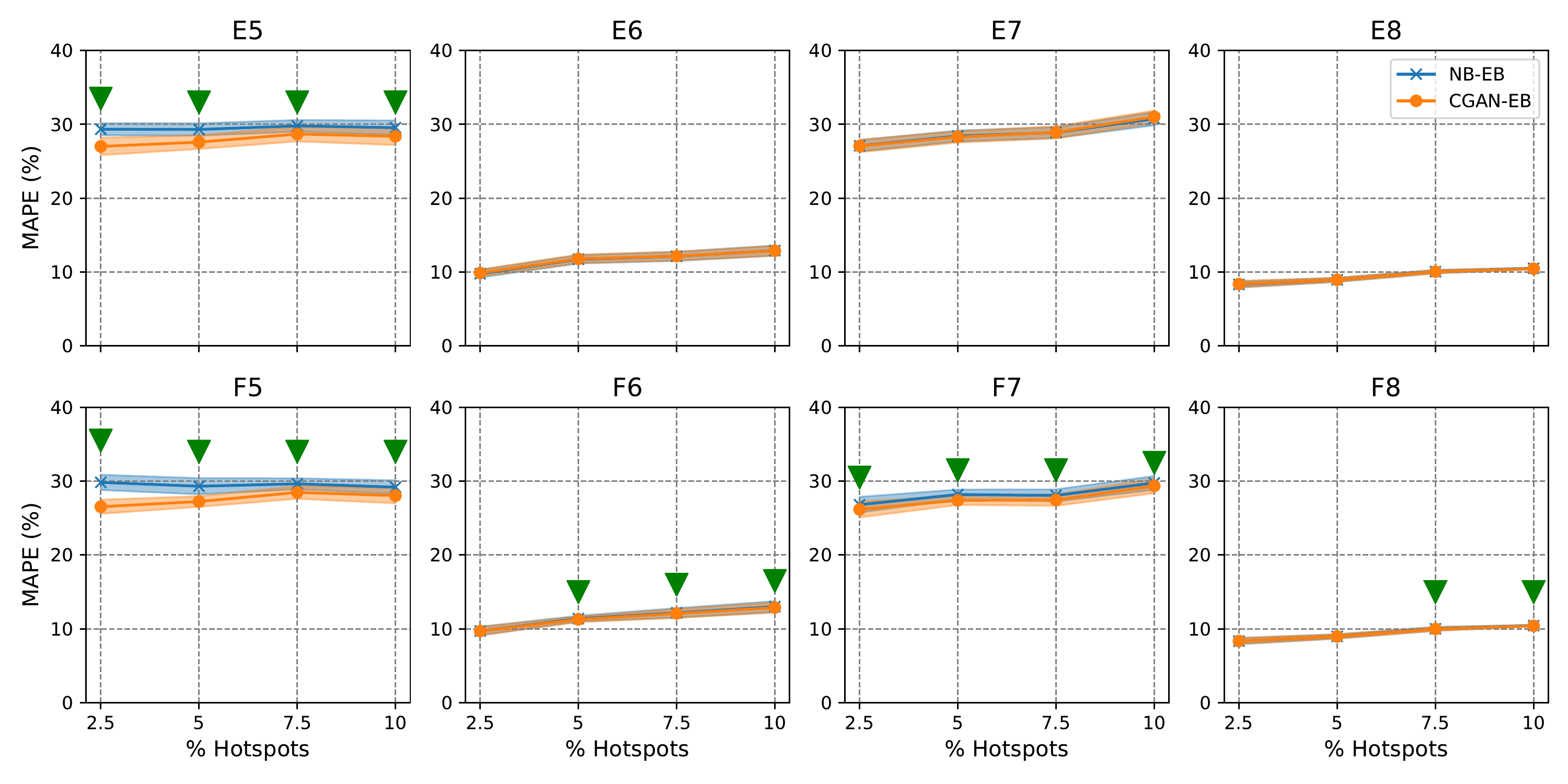}
    \caption{MAPE results for E-experiments versus F-experiments}
    \label{fig:MAPE}
\end{figure}

In order to compare the performance of CGAN-EB and NB-EB in terms of the estimation accuracy for Poison mean in top hotspots, we found the 95\% confidence intervals for the mean absolute percentage error (MAPE) for E-experiments and F-experiments and, similar to the analysis of the FI and PMD tests, a paired sample t-test was performed for each experiment to check if the difference in MAPE from the two models is statistically significant. The MAPE confidence intervals are presented in Figure \ref{fig:MAPE} and cases for which the performance of the models are statistically different are indicated by the green triangle. The results indicate that CGAN-EB more accurately estimates the Poison mean (i.e. has lower MAPE) than the NB-EB for 4 of 16 cases in the E-experiments and 13 of 16 cases in the  F-experiments. The percentage improvement in MAPE error for CGAN-EB in E-experiments can reach up to 8\% (in E5) and up to 11\% (in F5) in F-experiments. Consistent with the FI and PMD results, the performance improvement of the CGAN-EB model is larger for the cases for which the sample mean is low (i.e. F5 and F7).

In summary, the results in this section suggest that when all conditions and assumptions are set in favor of NB-EB, CGAN-EB performs at least as good as NB-EB. However, when we change some of the simulation conditions such that they are not consistent with the assumptions of the NB model (such as the functional form) the difference between the two methods becomes more obvious and the CGAN-EB shows better performance especially when the sample mean is low.

\section{Conclusions and Recommendations}
\label{section: conclusion}

In this study, we have proposed a novel non-parametric empirical Bayes method based on conditional generative adversarial network (CGAN-EB) and compared its performance with the traditional parametric approach using negative binomial model (NB-EB) in a simulation environment. Several experiments have been conducted, each of which include simulating the data sets with defined parameters, fitting CGAN and NB models, and estimating EB estimates. Then the models are compared based on their ability to detect correct hotspots (using FI and PMD tests) as well as their accuracy of EB estimates (using MAPE measure). The  results show that despite the simulated data sets being  generated based on NB assumptions, the performance of CGAN-EB was at least as good as NB-EB performance. Furthermore, when the simulation conditions were altered so that the generated data did not entirely conform to the assumptions of the NB model, CGAN-EB performed better than NB-EB in terms of hotspot detection and EB estimation accuracy. In addition, CGAN-EB generally showed better performance in experiments with lower sample means which is a common characteristics of crash data sets \cite{lord2010crashdata}.  

These findings are important as they suggest the proposed CGAN-EB model is more robust than the traditional NB-EB model in that it is able to perform as well or better than the NB-EB model over a range of conditions.

Notwithstanding these highly promising results, a number of questions remain about the proposed approach that need to be examined in the future. 
\begin{enumerate}
\item In this paper the performance of CGAN-EB was evaluated in a simulation environment so that the truth is known. The next step is to evaluate its performance using real-world data sets and use available tests such as site consistency test, method consistency test, and rank difference test. 
\item The performance of CGAN is impacted by its configuration  (e.g. architecture, size). For all data sets in this paper, we used a simple network architecture; however, other types of architectures might be better options.  Consequently, it is recommended to examine the sensitivity of the CGAN-EB performance as a function of the CGAN configuration and to determine if different configuration are better suited for different types of network screening applications. 
\item  Temporal and spatial transferability of SPFs is an area of concern for road safety professionals. The current study has shown than the proposed CGAN-EB model is robust in that it performs as well or better than the NB-EB model, largely because the CGAN model can automatically adapt to changes in the underlying crash data.  We hypothesize that these same characteristics will enable to the proposed CGAN-EB model to have greater transferability than conventional parametric models, including the NB-EB model.  However, this hypothesis needs to be tested. 
\item  There are other approaches than NB-EB for network screening such as finite mixture or zero-inflated models, which have been shown to be better alternatives in some conditions \cite{park2014finite, sharma2013zero}. It is recommended to compare the performance of CGAN-EB with such alternatives for crash data sets for which these models are more appropriate than the NB-EB.
\item Finally, studies have shown that using hierarchical full Bayesian method outperforms the standard EB approach in correctly identifying hazardous sites \cite{huang2009empirical, persaud2010comparison}. Future studies can focus on investigating how to combine this method with a non-parametric model such as CGAN model in order to improve the network screening performance. 

\end{enumerate}






\bibliographystyle{Other/model1-num-names.bst}
\bibliography{main.bib}







\end{document}